%% file: main.tex
\documentclass[runningheads]{llncs}
\usepackage[table]{xcolor}
\usepackage{multirow}
\usepackage{caption} 
\usepackage{subcaption}
\usepackage{graphicx}

\usepackage{scalerel}
\usepackage{tikz}
\usetikzlibrary{svg.path}

\usepackage{comment}
\usepackage{amsmath,amssymb} 
\usepackage{color}
\usepackage{soul}

\definecolor{lightgraytable}{HTML}{EFEFEF}
\DeclareRobustCommand{\hlgrey}[1]{{\sethlcolor{lightgraytable}\hl{#1}}}

\usepackage[accsupp]{axessibility}  

\usepackage{cite}
\usepackage{pifont}
\usepackage[pagebackref=true,breaklinks=true,colorlinks,bookmarks=false]{hyperref}

\begin{document}
\pagestyle{headings}
\mainmatter
\def\ECCVSubNumber{18}  

\def\etal{\emph{et al.}}

\title{MUST-VQA: MUltilingual Scene-text VQA} 

\titlerunning{MUST-VQA}
%
\author{Emanuele Vivoli\inst{1, 2} \and
Ali Furkan Biten\inst{2} \and
Andres Mafla\inst{2} \and
\\ Dimosthenis Karatzas\inst{2} \and
Lluis Gomez\inst{2}}

\authorrunning{E. Vivoli et al.}
%
\institute{University of Florence, Italy\\
\email{emanuele.vivoli@unifi.it}\and
Computer Vision Center, UAB, Barcelona, Spain\\
\email{\{abiten, amafla, dimos, lgomez\}@cvc.uab.es}}
\maketitle

\begin{abstract}
In this paper, we present a framework for Multilingual Scene Text Visual Question Answering that deals with new languages in a zero-shot fashion.
Specifically, we consider the task of Scene Text Visual Question Answering (STVQA) in which the question can be asked in different languages and it is not necessarily aligned to the scene text language.
Thus, we first introduce a natural step towards a more generalized version of STVQA: MUST-VQA. Accounting for this, we discuss two evaluation scenarios in the constrained setting, namely IID and zero-shot and we demonstrate that the models can perform on a par on a zero-shot setting. We further provide extensive experimentation and show the effectiveness of adapting multilingual language models into STVQA tasks. 

\end{abstract}

%
%


%
%

\section{Introduction}




Visual Question Answering is a prominent task that involves two modalities: vision and language. Language is not only used for expressing the question to the model, but it's sometimes implicit in the context of text found in the image, 
such as in the case of Scene Text Visual Question Answering (STVQA) task~\cite{biten2019scene,singh2019towards}. 
The ultimate goal for a holistic STVQA model is to be able to accept questions, read/analyze the scene text and produce answers in any language or script, this scenario is referred as \textit{unconstrained setting}. 
This is especially true considering the fact that there currently exist more than 7k spoken languages, while more than 4k have a developed writing system\footnote{https://www.ethnologue.com/enterprise-faq/how-many-languages-world-are-unwritten-0}, spanning over 100 different scripts.
We believe that the natural extension of the STVQA task in order to benefit more people while reaching a wider use case, it has to have the capabilities of dealing with MUltilingual STVQA (MUST-VQA).

Evidently, reaching this goal is far from easy as it encapsulates dealing with multiple problems. One of the most important problems is the data scarcity in questions as well as finding images that contain scene text in various languages, being particularly difficult in low-resource languages. Therefore, it is infeasible to collect data for all the languages with all the possible scripts. Moreover, even though STVQA has attracted a lot of research~\cite{biten2019icdar,gomez2021multimodal,hu2020iterative,yang2021tap,biten2022latr}, the dataset in itself is designed solely for English text. This significantly underpins its use and application in a practical manner considering that roughly 80\% of the world population does not speak English~\cite{crystal2008two}. Given the difficulties of obtaining new data and having only English readily available dataset, we define a new practical \textit{constrained setting}. In this setting, we assume that we have questions in multiple languages apart from English. We further divide the constrained setting into IID and zero-shot setting where the models are evaluated with the languages the model is trained with and the languages the models have never seen before, respectively. The zero-shot setting allows models to extend to low-resource languages.
Thus, the constrained setting acts as the first step towards the unconstrained, and our aim is to study the behaviour of various models with questions asked in languages other than English. 


More specifically, in this work, we take the first steps towards MUltilingual STVQA (MUST-VQA) by automatically translating all the questions in ST-VQA~\cite{biten2019scene} and TextVQA~\cite{singh2019towards} to 5 languages with 3 scripts; namely Spanish, Catalan, Chinese, Italian and Greek, by using automatic translation models and evaluate on IID and zero-shot settings. Furthermore, it is known that neural networks are prone to exploiting shortcuts~\cite{geirhos2020shortcut} and thus, we examine our models' robustness to distinct machine translation models. 
Finally, we study the effect of multiple STVQA models and possible ways to adapt the original architectures to incorporate multilingual inputs.

Our work aims at finding the limitations of the models in MUST-VQA as a preceding step before tackling a full unconstrained multilingual setting. 
The main contributions of our work are:
\begin{itemize}
    \item We introduce a natural step towards a more generalized version of STVQA, MUST-VQA and define two settings, unconstrained and constrained. 
    
    \item We discuss two evaluation scenarios in the constrained setting, namely IID and zero-shot and we demonstrate that our proposed models can perform at the same level in a zero-shot setting.
    
    \item We provide extensive experimentation and show the effectiveness of adapting multilingual language models into STVQA tasks. 

        
                
\end{itemize}

%
%

\section{Related work}

The use of scene text in the VQA task is a recent trend in vision and language models' research. Many datasets have been published considering scene text in different domains: natural images~\cite{biten2019scene,singh2019towards}, scanned documents~\cite{mathew2021docvqa} book and movie covers~\cite{mishra2019ocr}, and info-graphics~\cite{mathew2021infographicvqa}.
Additionally, a bilingual (English+Chinese) dataset  has been proposed for VQA~\cite{han2020finding}, as well as a captioning dataset with natural images~\cite{sidorov2020textcaps}.
%

Alongside with all these datasets, state of the art models have evolved significantly. Singh~\etal~\cite{singh2019towards} introduced a pointer network to answer either with an answer from the fixed answer vocabulary or by selecting one of the OCR strings. Gómez~\etal~\cite{gomez2021multimodal} also employed pointer networks directly to the image pixels instead of selecting from a vocabulary. 
Hu~\etal~\cite{hu2020iterative} as well used pointer networks 
with a single multi-modal transformer (M4C) to encode all modalities together.
Kant~\etal~\cite{kant2020spatially} built on top of M4C with a spatially aware self-attention layer such that each visual entity only looks at neighboring entities defined by a spatial graph.
Zhu~\etal~\cite{zhu2020simple} proposes to use an attention mechanism to fuse pairwise modalities. 

Recently, following its success in language models~\cite{devlin2018bert,liu2019roberta}, pre-training has also been successfully used in STVQA. Yang~\etal~\cite{yang2021tap} performed two stage training where first they do pre-training in a large corpus of images with text to conduct several pretext tasks (OCR token relative position prediction,  masked language modelling, and image-text matching) and later fine-tuning for the STVQA task, showing huge performance gains.  
Finally, Biten~\etal~\cite{biten2022latr} used layout information via pre-training on IDL~\cite{biten2022ocr} data to achieve state-of-the-art performance across multiple benchmarks. 

However, the main assumption made until now is that the language of the \textit{question}, \textit{text in the image} and \textit{answer} is always English.
Our belief is that the task of MUST-VQA is still unexplored and lack robust benchmarks. Some recent work has approached the problem of Multilingual Scene-Text VQA (MUST-VQA), but their studies were limited to the use of mono-lingual models (one model per language)~\cite{han2020finding}, or to a single old-fashioned VQA architecture~\cite{brugues2022multilingual}.


In this work, we  define a customized version of two state of the art transformer-based STVQA models (M4C~\cite{hu2020iterative} and LaTr~\cite{biten2022latr}) to incorporate multi-lingual inputs in a constrained scenario. We employ both approaches as benchmarks for the proposed MUST-VQA task.



%
%

\section{Method}
In this section the main building blocks of our models are introduced and explained. 
We start by formally defining the task of MUST-VQA in the constrained and unconstrained settings, and then we describe each of these modules.

\subsection{Task Definition}

Let $v \in I$ be an image of the image space $I$, and $q \in Q$ a question belonging to the question space $Q$.  The ultimate goal for VQA is to be able to accept questions $q$ and an image $v$ to produce an answer $a \in A_{v,q}$. In our case, we focus on STVQA, task in which the image $v \in \widetilde{I}$ contains scene text, and the question $q \in \widetilde{Q}$ is related to the text in the image. However, the actual state-of-the-art is not able to handle an unconstrained setting, since current models are only trained in English. Therefore, we define additional elements that help towards our goal. First, let $\widetilde{I}_{en} \subset I$ be the subspace of Images containing English scene text, and $\widetilde{Q}_{en} \subset Q$ be the subspace of English questions about text in the image. Let be $OCR_{sys}$ a blackbox which takes as input an image $v \in \widetilde{I}_{en}$ and outputs a set $T = \{ (t_{v}^{i}, b_{v}^{i}) | i = 0, 1, \dots \}$ where $t_{u}^{i}$ is a token, and $b_{u}^{i} \in [0,1]^4$ is its normalized position in the image. A common STVQA architecture is able to process all these modalities $v \in \widetilde{I}_{en}$, $q \in \widetilde{Q}_{en}$, $T = OCR_{sys}(v)$ and produce an answer. In order to do that, we need to define some architecture modules. 

\begin{figure}
    \centering
    \includegraphics[scale=0.30]{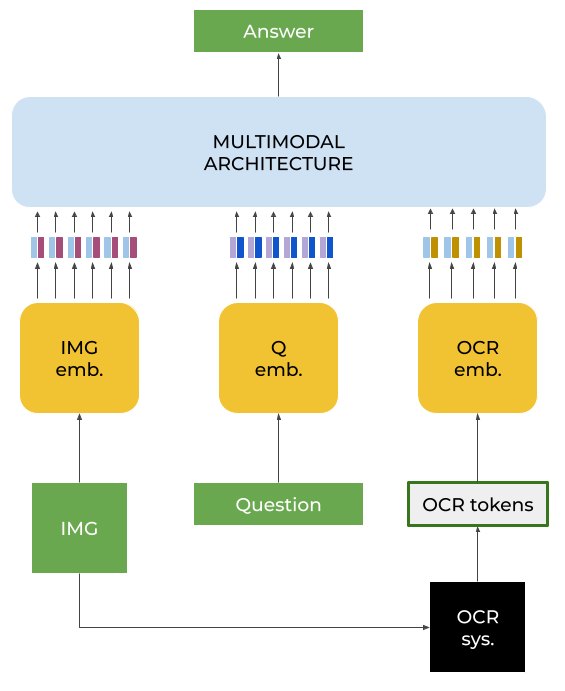}
    \caption{Proposed Model}
    \label{fig:STVQA-Arch}
\end{figure}

As we can see from Figure \ref{fig:STVQA-Arch}, given an image $v \in \widetilde{I}_{en}$ we obtain a set of M visual features $x_{vis}^{m}$ and positional information $x_{b}^{m}$ through an $IMG_{emb}$ as $\{ ( x_{vis}^{m}, x_{b}^{m} ) | m = 1, \dots M\} = IMG_{emb}(v)$. Additionally, given the question $q \in \widetilde{Q}_{en}$ and a module $Q_{emb}$, we obtain a set of N textual features $y_{q}^{n}$ and positional information $y_{b}^{n}$ as $\{ ( y_{q}^{n}, y_{b}^{n} ) | n = 1, \dots N\} = Q_{emb}(q)$. Lastly, taking into consideration the remaining modality, which is OCR tokens, we can obtain a set of $|T|$ textual features as $ z_{u}^{i} = OCR{emb}(t_{u}^{i})$ with $i = 0 \dots |T|$.

\noindent\textbf{MUST-VQA.} Until now, the set $\widetilde{I}_{en}$ and $\widetilde{Q}_{en}$ have been defined as set of Images containing scene text and set of Questions about text in the images. However, in the common STVQA task a strong bias is added to the selection of these subsets of $I$, $Q$ and $A$: the language. In fact, in STVQA the three elements which are question, scene text and answer all have in common the English language. Thus, we can sample the subspaces $\widetilde{I}_{es} \subset \widetilde{I}$ to get images containing text in Spanish, as well as sample the subspace $\widetilde{Q}_{zh} \subset \widetilde{Q}$ to get Chinese questions about text in the image. The same is true also for the set of answers. With that said, the unconstrained setting of MUST-VQA covers using any sampling, with respect to the language, from $\widetilde{I}$ and $\widetilde{Q}$. However, for most language combinations in the world, data availability is limited, which makes it difficult to obtain for example images with Spanish scene text and original Chinese questions. To this end, we define the constrained MUST-VQA task in which multiple question sets are generated from $\widetilde{Q}_{en}$ by means of an external translator module. The question sets generated with translator $g$ are referred to as $\widetilde{Q}_{ca}^g, \widetilde{Q}_{es}^g, \widetilde{Q}_{zh}^g$, etc. By doing this, we define two experimental settings: IID, in which a subset of question sets is used for training and testing a multimodal architecture $\{ \widetilde{Q}_{l}^g | l \in (en, ca, es, zh) \}$, and Zero-shot in which we want to test the language transfer capabilities of our models trained under the IID setting to a subset of other languages $\{ \widetilde{Q}_{l}^g | l \in (it, el) \}$.

\subsection{Visual Encoder}
In order to obtain the most salient regions of a given image, a Faster R-CNN~\cite{ren2015faster} is used, as proposed initially by~\cite{anderson2018bottom} and employed in STVQA models as in~\cite{singh2019towards, yang2021tap}. The employed Faster R-CNN is pre-trained in ImageNet~\cite{russakovsky2015imagenet}. Later, Visual Genome~\cite{krishna2017visual} is employed to fine-tune the Faster R-CNN to not only predict classes, but also incorporate the prediction of attributes that belong to a specific region that contains an object. 
The resulting model is employed to extract a set of bounding boxes and visual features enclosed in such regions. In all of our models, the features obtained are then passed through a trainable linear projection layer. The resulting visual features are later fed to each explored model.


\subsection{Textual Encoders}

In this section, we describe the different textual encoders that have been employed to obtain language features. Specifically, we embed the questions through a given encoder to obtain a set of features to be used as a representation to be later fed into a transformer-based model.
\\
\noindent\textbf{Byte Pair Encoding (BPEmb).} The BPEmb~\cite{heinzerling2017bpemb} is a variable-length encoding that treats text as a symbol sequence. It merges the most common pairs of symbols into a new representation in an iterative manner. This encoding method is trained on Wikipedia in a corpus that employs 275 different languages, thus creating a robust representation that includes most characters found in common human languages. It is shown experimentally that this approach yields rich representations of text that perform on a par compared to other subword embeddings such as Fasttext~\cite{bojanowski2017enriching}.
BPEmb does not require tokenization and is orders of magnitude smaller than alternative embeddings, allowing for potential applications, specially representing unseen words in different alphabets, thus making it a strong encoder in multilingual scenarios.
\\
\\
\noindent\textbf{Bidirectional Encoder Representations from
Transformers (BERT).} BERT~\cite{devlin2018bert} employs a multi-layer implementation based on the initial Transformer~\cite{vaswani2017attention}. The work from~\cite{devlin2018bert} incorporates two pre-training tasks. The first one, masked-language-modelling (MLM) focuses on predicting a masked tokenized word based on the surrounding words. This pretext task aims to learn semantic representations of words. The second pre-training task is next sentence prediction (NSP) which given a pair of sentence, the model has to predict whether these sentences are consecutive or not. BERT and variations inspired on it are commonly employed as strong semantic descriptors of text in natural language processing. However, the main drawback lies in the lack of sub-word processing to represent out of vocabulary words. 
\\
\\
\noindent\textbf{Multilingual-BERT (M-BERT).} As in BERT~\cite{devlin2018bert}, M-BERT is a 12 layer transformer, but rather than relying only on a monolingual English data corpus, it is trained on 104 Wikipedia sites on different languages that share a common vocabulary. 
It makes no use of a marker to indicate the input language, and there is no explicit mechanism in place to promote translation-equivalent pairings to have comparable representations. 
\\
\\
\noindent\textbf{Text-to-Text Transfer Transformer (T5).} The T5~\cite{raffel2020exploring} is an encoder-decoder transformer. Minor variations are employed from the original~\cite{vaswani2017attention} implementation.
The difference lies in that T5 employs a scaled-down form of layer normalization in which no additive bias is added and the activations are simply rescaled.
The T5 architecture is trained on the Colossal Clean Crawled Corpus (C4). The C4 is a text collection that is not only orders of magnitude larger than normal pre-training data sets (about 750 GB), but also comprises clean and curated English material. The model employs a similar query structure describing the task to be performed, such as translation, question answering and classification. The resulting approach can be applied to a variety of tasks, while at the same time similar loss function, model and hyper parameters can be used.
\\
\\
\noindent\textbf{Multilingual-T5.} The mT5~\cite{xue2020mt5} model employs a similar set of layers and design as T5. However, they differ in the training corpus. 
The mT5 model was trained on a 101-language Common Crawl-based multilingual variation. Only English Common Crawl is what T5 has been pre-trained on.
Additionally an increase in performance is obtained by the use of GeGLU nonlinearities~\cite{shazeer2020glu}.

\subsection{Baselines}\label{subsec:models}
In this section we introduce the Scene Text Visual Question Answering models adapted for MUST-VQA. First, we start by introducing the base-model details and then we describe the customized modifications performed on each of them to better adjust to handle multilingual inputs.

\subsubsection{M4C} Multimodal Multi-Copy Mesh (M4C)~\cite{hu2020iterative} is a multimodal transformer architecture that employs a dynamic pointer network to select among a fixed dictionary or scene text instances. The input comes from two modalities, question and image. However, a scene text recognizer is employed to extract textual instances, which also serve as input to the M4C model. The questions are encoded using BERT~\cite{devlin2018bert}, while a list of visual object features are obtained by using an off-the-shelf object detector Faster R-CNN~\cite{ren2015faster}. The scene text tokens are obtained by relying on an OCR module, Rosetta-en~\cite{borisyuk2018rosetta}. The resulting textual transcription is embedded with FastText~\cite{bojanowski2017enriching} and a Pyramidal Histogram Of Characters (PHOC)~\cite{almazan2014word}. Such embeddings have shown to be robust representations to encode semantics and morphological information of text~\cite{almazan2014word, mafla2020fine}. The resulting embedded scene text is projected to the same dimension as all other text and visual tokens in order to be used as input to a transformer. The answers are produced in an iterative manner, while the model either selects to output words from a fixed vocabulary or from OCR tokens found in an image by employing a dynamic pointer network.

\subsubsection{M5C} The proposed Multilingual-M4C (M5) underwent through a set of custom modifications in order to be able to accept different languages aside from English. 
To accomplish this goal, we designed a new model:
\textit{M5C-mBERT}. The first modification is to substitute the FastText embedding of OCR tokens, since FastText is pre-trained only on English. Next, we replaced the PHOC representation to be able to incorporate different scripts. PHOC encodes only latin-based scripts, therefore it is not suitable for handling unknown languages unless a big descriptor is employed. Therefore we employed a multi-language aligned text embedding method such as BPEmb~\cite{heinzerling2017bpemb}. In this baseline, we introduce a multilingual Language Model for the question embedding, instead of a pre-trained English based BERT, thus lacking the capability of embedding different languages. By doing that, we designed M5C-mBERT to have multilingual BERT for question embedding.

\subsubsection{LaTr (T5)}  In~\cite{biten2022latr}, a Layout-Aware Transformer (LaTr) is proposed, which is based on a T5~\cite{raffel2020exploring} encoder-decoder architecture. The pipeline consists of three modules. The first one consists of a Language Model trained specifically on document layouts~\cite{biten2022latr} which contains only text and layout information. The second module is an spatial embedding designed to embed scene text tokens along with positional information. Lastly, a ViT~\cite{dosovitskiy2020image} is employed to extract visual tokens. All these three modalities are employed as input to the pre-trained transformer. The encoder learns a suitable representation of the alignment of the 3 modalities to later be used by a decoder to reason and output an answer.

\subsubsection{mLaTr (mT5)} In this baseline, we replaced the T5 encoder-decoder transformer with the mT5 model in LaTr. Differently from LaTr (which uses layout aware pre-training), we fine-tuned only the text pre-trained multi-lingual Language Model with the multimodal information. Therefore, the input to this mT5 transformers are questions tokens, OCR tokens, and visual features.

\section{Experiments}
We consider the standard benchmarks of ST-VQA~\cite{biten2019scene} and TextVQA~\cite{singh2019towards}. The proposed MUST-VQA datasets consists of ML-STVQA and ML-TextVQA which are obtained by translating ST-VQA and TextVQA into \textit{Catalan}, \textit{Spanish}, \textit{Chinese}, \textit{Italian}, and \textit{Greek} with Google-Translate-API\footnote{cloud.google.com/translate/}, resulting in a multi-lingual datasets comprised by 6 languages for the constrained setting task of MUST-VQA.
In this section, we experimentally examine our baselines in the constrained setting. We further test these baselines for zero-shot multilingual setting, both on ML-STVQA and ML-TextVQA datasets. 

\subsection{Implementation Details}\label{subsec:imp-details}





For all M4C-based methods we used Adam~\cite{kingma2014adam} optimizer, with learning rate of 1e-4, and a learning rate decreasing at 14k and 19k iterations. The final model is trained for 24k, while using a 128 batch size. 

For all T5-based models we used AdamW~\cite{loshchilov2017decoupled} optimizer with learning rate of 1e-4, employing a warm up for the first 1000 iterations until reaching 1e-3. Afterwards, we decreased to zero linearly until the end of the training. The batch size employed was 128 and models were trained for 24k iterations for the ML-STVQA dataset. The model trained on ML-TextVQA dataset employed 48k iterations and a batch of 64. 


\subsection{TextVQA Results}

In this section we evaluate results for the dataset ML-TextVQA. We define two evaluation settings. The former is the constrained setting that only uses English, Catalan, Spanish, and Chinese questions for training. On these languages, all the models presented in Section \ref{subsec:models} are trained following the config specifications in \ref{subsec:imp-details}. Here, either Rosetta-OCR or Microsoft-OCR are used for detection. 
The latter is the zero-shot transfer setting, in which we measure the performance of the previous models on two new languages that the model has not seen during training (Italian and Greek).
\input{tables/textvqa-iid}

\label{subsec:TextVQA-IID}
\noindent\textbf{IID Languages (en, ca, es, zh).} The first part of Table \ref{tab:ML_TextVQA-iid} presents training using Rosetta OCR System, while the bottom part using Microsoft-OCR.
Background color is employed to distinguish between monolingual models (white) and our multilingual models (\hlgrey{grey}). 
As can be appreciated, our multilingual \textit{M5C-mbert} outperforms \textit{M4C} of about \textbf{+1.71\%} and \textbf{+2.44\%} with Rosetta-OCR and Microsoft-OCR respectively, with fewer parameters. These values are the average over the four languages calculated by combining all four subset into a single one. Moreover, as a multilingual model, it is able to perform on Chinese \textbf{+5.94\%} and \textbf{+8.75\%} better than its English counterpart (M4C). Increasing model capability to \textit{mLaTr-base} results in a performance gain of \textbf{+3.8\%}. Furthermore, when training using visual features, performances either recorded a loss of -0.03\% and -0.11\% for \textit{LaTr-base} Rosetta-OCR and \textit{mLaTr-base} Microsoft-OCR or an increase of +0.58\% and +0.16\% for \textit{mLaTr-base} Rosetta-OCR and \textit{LaTr-base} Microsoft-OCR. Thus, performances difference is very marginal. Finaly, from \textit{LaTr-base} with Microsoft-OCR and visual features we notice that it obtain the best accuracy on Validation set for English Language, which might be due to the distribution of pre-training only-english data samples. In fact, \textit{T5} model has been trained on huge amount of English transcripts (C4), which consist on cleaned English texts from Common Crawl.

\input{tables/textvqa-zero_shot}

\label{subsec:ML_TextVQA-zero_shot}
\noindent\textbf{Zero-shot Transfer (it, el).} A more challenging case for MUST-VQA is the zero-shot cross-lingual setting. Here, a pretrained multilingual model is fine-tuned on TextVQA considering a set of languages but tested on others. In our constrained setting this means testing the models in Section \ref{subsec:models} to generate English answers from Italian or Greek questions, despite having only seen English, Catalan, Spanish and Chinese questions during training.
A note for Table \ref{tab:ML_TextVQA-zero_shot}: last column \textit{Avg.} is the accuracy calculated by combining all four IID subset into a single one. A major observation can be made from Table \ref{tab:ML_TextVQA-zero_shot}: the best model for IID setting, also perform better on the task of Zero-shot transfer to unseen languages. Moreover, while for Italian the difference is tangible (\textbf{+7.92\%}), for Greek the gap becomes even wider (\textbf{+22.77\%}). This behavior might have two main reasons: (1) Italian, Catalan and Spanish are part of the Roman family, descended from Latin, while Greek does not have this common roots with them \cite{DBLP:journals/corr/abs-2001-11899}; (2) Italian share the same script with English, Catalan, and Spanish while Greek has its own script. From these facts we can justify that English-only models trained under constrained settings of EN, CA, ES, ZH languages do have the linguistic and scripting capability of transfer knowledge to Italian setting resulting in \textbf{37.08\%} accuracy at best, but do not have the same potential for Greek.

\input{tables/stvqa-iid}

\subsection{ST-VQA Results}

\noindent\textbf{IID Languages (en, ca, es, zh).} Table \ref{tab:ML_STVQA-iid} presents the Accuracy and ANLS values in the constrained and unconstrained settings. Similarly to section \ref{subsec:TextVQA-IID}, the upper part of the Table refers to Rosetta-OCR, while the bottom to Microsoft-OCR. The \hlgrey{grey} lines indicates multilingual models, while the white English-only models. However, all these models have been trained on ML-STVQA. One thing to notice is that in this dataset, the best performance is obtained by the \textit{mLaTr-base} model with Microsoft-OCR and visual features. With that said, Chinese is the only exception in which the \textit{mLaTr-base} configuration without visual features actually performs slightly better if considering Accuracy itself. Thus, this empirically confirms, also for this dataset, the fact that visual features might not be relevant to this task. Regarding the comparison of different models in the same ML-STVQA dataset, we can notice once more that \textit{M5C-mbert} obtained \textbf{+1.77\%} and (+1.65\%) increase in terms of accuracy with respect to M4C English-only baseline. Moreover, from Table \ref{tab:ML_STVQA-iid}, we can appreciate three main facts: (1) \textit{mLaTr-base} obtains the best result in overall accuracy, in its variation using Microsoft-OCR and visual features. However, we also observe that visual features don't have considerable impact on the results. (2) When focusing on each language, in the upper part of the table (with Rosetta-OCR) results show that even if \textit{LaTr-base} English-only performs worse than \textit{mLaTr-base} multilingual on almost all the languages with the bigger margin of \textbf{-15.72\%} for Chinese, it still outperforms the multilingual version for English questions by almost 1 point (\textbf{+0.95\%}). The last consideration (3) is regarding the pointer network against generative models for languages out of vocabulary. In fact, despite having the lowest score in the overall results, \textit{M4C} obtains higher accuracy in the Chinese questions with both OCR systems, resulting in a margin of \textbf{+6.05\%} (Rosetta-OCR) and \textbf{+7.61\%} (Microsoft-OCR) compared to \textit{LaTr-base}.

\input{tables/stvqa-zero_shot}

\noindent\textbf{Zero-shot Transfer (it, el).} From Table \ref{tab:ML_STVQA-zero_shot} we can see that the best model for IID setting, also performs better on the task of Zero-shot transfer to unseen languages. Moreover, as saw for ML-TextVQA zero-shot, while for Italian the difference is tangible (+5.82\%), for Greek the gap becomes even wider (+17.81\%). Possible reasons for that are commented in \ref{subsec:ML_TextVQA-zero_shot}.





\section{Analysis}






\noindent\textbf{Robustness to Translation models} In our method, in order to obtain questions in different languages, a translation model is used. Our original translation model is Google-Translate, accessed from its API. To study our approach and how a translation model can influence results, we use three other machine translation models, namely OPUS, M2M\_100 and mBART. For all these translation models, we calculate the accuracy of our best model (\textit{mLaTr-base}) for different languages, in term of IID and Zero-shot settings. From Table \ref{tab:translate-robust} we can see that accuracy does not drop with other translation models, but instead it has values coherent with the original translation model we use. 

\begin{table}[!htb]
    \tiny
    \caption{Results refer to \textit{mLaTr-base} with visual features and Microsoft-OCR. Its average accuracies on Original ML-TextVQA and ML-STVQA questions are reported in the last column \textit{Avg}. Questions have been translated into the 5 languages using OPUS, M2M100 (1.2B), and mBART.}
    \begin{subtable}{.5\linewidth}
      \centering
        \caption{\textbf{Results on TextVQA dataset}}
        \input{tables/textvqa_translate_robustness-tabular}
    \end{subtable}%
    \begin{subtable}{.5\linewidth}
      \centering
        \caption{\textbf{Results on STVQA dataset}}
        \input{tables/stvqa_translate_robustness-tabular}
    \end{subtable} 
    \label{tab:translate-robust}
\end{table}








\section{Conclusions and Future Work}
In this paper, we present a framework for Multilingual visual question answering that deals with new languages in a zero-shot fashion.
Specifically, we defined the task of MUST-VQA and its constrained and unconstrained settings. We defined a multilingual baseline method for MUST-VQA by adopting monolingual architectures. Our results suggest that it is able to operate in a zero-shot fashion, and independent on the translation method used to obtain multilingual questions.
In this work, the constrained setting acts as the first step towards the unconstrained, and our aim is to study the behaviour of various models with questions asked in languages other than English. Further work will need to approach also answers in different languages, probably matching the question language.

\subsection*{Acknowledgments}
This work has been supported by projects PDC2021‐121512‐I00, PLEC2021‐00785, PID2020-116298GB-I00, ACE034/21/000084, the CERCA Programme / Generalitat de Catalunya, AGAUR project 2019PROD00090 (BeARS), the Ramon y Cajal  RYC2020-030777-I / AEI / 10.13039/501100011033 and PhD scholarship from UAB (B18P0073).

\clearpage
%
%
\bibliographystyle{splncs04}
\bibliography{egbib}

\end{document}

%% file: tables/textvqa-iid.tex
\begin{table}[t!]
\centering
\begin{tabular}{llccllll|l}
\hline
\rowcolor[HTML]{FFFFFF} 
Method     & OCR    & Vis. Feat. & Params                      & \multicolumn{1}{c}{\cellcolor[HTML]{FFFFFF}EN}                & \multicolumn{1}{c}{\cellcolor[HTML]{FFFFFF}CA}                & \multicolumn{1}{c}{\cellcolor[HTML]{FFFFFF}ES}                & \multicolumn{1}{c|}{\cellcolor[HTML]{FFFFFF}ZH}               & \cellcolor[HTML]{FFFFFF}Avg.  \\ \hline
\rowcolor[HTML]{FFFFFF} 
M4C        & Ros-en & \ding{52}          & 200M                         & \cellcolor[HTML]{FFFFFF}{\color[HTML]{333333} 28.96}          & {\color[HTML]{333333} 29.9}                                   & {\color[HTML]{333333} 29.60}                                  & {\color[HTML]{333333} 23.73}                                  & \cellcolor[HTML]{FFFFFF}28.44 \\
\rowcolor[HTML]{EFEFEF} 
M5C-mbert  & Ros-en & \ding{52}          & {\color[HTML]{333333} 162M} & {\color[HTML]{333333} 28.83}                                  & \cellcolor[HTML]{EFEFEF}{\color[HTML]{333333} 30.26}          & \cellcolor[HTML]{EFEFEF}{\color[HTML]{333333} 30.35}          & {\color[HTML]{333333} 29.67}                                  & 30.15                         \\
\rowcolor[HTML]{FFFFFF} 
LaTr-base  & Ros-en & \ding{55}          & 226M                        & \cellcolor[HTML]{FFFFFF}{\color[HTML]{333333} 41.02}          & {\color[HTML]{333333} 38.35}                                  & {\color[HTML]{333333} 38.94}                                  & {\color[HTML]{333333} 20.24}                                  & \cellcolor[HTML]{FFFFFF}34.64 \\
\rowcolor[HTML]{EFEFEF} 
mLaTr-base & Ros-en & \ding{55}          & 586M                        & {\color[HTML]{333333} 40.35}                                  & \cellcolor[HTML]{EFEFEF}{\color[HTML]{333333} 39.50}          & \cellcolor[HTML]{EFEFEF}{\color[HTML]{333333} 39.70}          & \cellcolor[HTML]{EFEFEF}{\color[HTML]{333333} 39.49}          & 39.77                         \\
\rowcolor[HTML]{FFFFFF} 
LaTr-base  & Ros-en & \ding{52}          & 226M                        & \cellcolor[HTML]{FFFFFF}{\color[HTML]{333333} 40.92}          & {\color[HTML]{333333} 38.40}                                  & {\color[HTML]{333333} 38.81}                                  & {\color[HTML]{333333} 20.34}                                  & \cellcolor[HTML]{FFFFFF}34.61 \\
\rowcolor[HTML]{EFEFEF} 
mLaTr-base & Ros-en & \ding{52}          & 586M                        & {\color[HTML]{333333} 40.96}                                  & \cellcolor[HTML]{EFEFEF}{\color[HTML]{333333} 40.35}          & \cellcolor[HTML]{EFEFEF}{\color[HTML]{333333} 40.35}          & \cellcolor[HTML]{EFEFEF}{\color[HTML]{333333} 39.78}          & 40.35                         \\ \hline
\rowcolor[HTML]{FFFFFF} 
M4C        & Ms-OCR & \ding{52}          & 200M                         & \cellcolor[HTML]{FFFFFF}{\color[HTML]{333333} 42.16}          & {\color[HTML]{333333} 41.89}                                  & {\color[HTML]{333333} 41.64}                                  & {\color[HTML]{333333} 33.60}                                  & \cellcolor[HTML]{FFFFFF}39.82 \\
\rowcolor[HTML]{EFEFEF} 
M5C-mbert  & Ms-OCR & \ding{52}          & 162M                        & {\color[HTML]{333333} 42.36}                                  & \cellcolor[HTML]{EFEFEF}{\color[HTML]{333333} 42.15}          & \cellcolor[HTML]{EFEFEF}{\color[HTML]{333333} 42.14}          & \cellcolor[HTML]{EFEFEF}{\color[HTML]{333333} 42.35}          & 42.26                         \\
\rowcolor[HTML]{FFFFFF} 
LaTr-base  & Ms-OCR & \ding{55}          & 226M                        & \cellcolor[HTML]{FFFFFF}{\color[HTML]{333333} 46.93}          & {\color[HTML]{333333} 44.32}                                  & {\color[HTML]{333333} 44.87}                                  & {\color[HTML]{333333} 23.18}                                  & \cellcolor[HTML]{FFFFFF}39.83 \\
\rowcolor[HTML]{EFEFEF} 
mLaTr-base & Ms-OCR & \ding{55}          & 586M                        & {\color[HTML]{333333} 46.63}                                  & \cellcolor[HTML]{EFEFEF}{\color[HTML]{333333} \textbf{46.10}} & \cellcolor[HTML]{EFEFEF}{\color[HTML]{333333} \textbf{46.12}} & \cellcolor[HTML]{EFEFEF}{\color[HTML]{333333} 45.38}          & \textbf{46.06}                         \\
\rowcolor[HTML]{FFFFFF} 
LaTr-base  & Ms-OCR & \ding{52}          & 226M                        & \cellcolor[HTML]{FFFFFF}{\color[HTML]{333333} \textbf{47.25}} & {\color[HTML]{333333} 44.15}                                  & {\color[HTML]{333333} 44.81}                                  & {\color[HTML]{333333} 23.79}                                  & \cellcolor[HTML]{FFFFFF}39.99 \\
\rowcolor[HTML]{EFEFEF} 
mLaTr-base & Ms-OCR & \ding{52}          & 586M                        & {\color[HTML]{333333} 46.65}                                  & \cellcolor[HTML]{EFEFEF}{\color[HTML]{333333} \textbf{46.09}} & \cellcolor[HTML]{EFEFEF}{\color[HTML]{333333} 45.58}          & \cellcolor[HTML]{EFEFEF}{\color[HTML]{333333} \textbf{45.44}} & \textbf{45.95}
\end{tabular}
\caption{\textbf{Results on the ML-TextVQA dataset}. Results refer to multi-lingual training on English, Catalan, Spanish, and Chinese and are reported in term of Accuracy.
}
\label{tab:ML_TextVQA-iid}
\end{table}


%% file: tables/textvqa-zero_shot.tex
\begin{table}[t!]
\centering
\begin{tabular}{llccll|l}
\hline
\rowcolor[HTML]{FFFFFF} 
Method     & OCR    & Vis. Feat. & Params                      & \multicolumn{1}{c}{\cellcolor[HTML]{FFFFFF}IT} & \multicolumn{1}{c|}{\cellcolor[HTML]{FFFFFF}EL} & Avg.  \\ \cline{1-7}
\rowcolor[HTML]{FFFFFF} 
M4C        & Ros-en & \ding{52}          & 200M                         & {\color[HTML]{333333} 17.45}                  & {\color[HTML]{333333} 5.84}                   & 28.44 \\
\rowcolor[HTML]{EFEFEF} 
M5C-mbert  & Ros-en & \ding{52}          & {\color[HTML]{333333} 162M} & {\color[HTML]{333333}24.92}                  & {\color[HTML]{333333}10.88}                   & 30.15 \\
\rowcolor[HTML]{FFFFFF} 
LaTr-base  & Ros-en & \ding{55}          & 226M                        & {\color[HTML]{333333}33.35}                  & {\color[HTML]{333333}18.02}                   & 34.64 \\
\rowcolor[HTML]{EFEFEF} 
mLaTr-base & Ros-en & \ding{55}          & 586M                        & {\color[HTML]{333333}3873}                  & {\color[HTML]{333333}37.78}                   & 39.77 \\
\rowcolor[HTML]{FFFFFF} 
LaTr-base  & Ros-en & \ding{52}          & 226M                        & {\color[HTML]{333333} 33.59}                   & {\color[HTML]{333333} 15.01}                    & 34.61 \\
\rowcolor[HTML]{EFEFEF} 
mLaTr-base & Ros-en & \ding{52}          & 586M                        & {\color[HTML]{333333} 39.45}                   & {\color[HTML]{333333} 38.03}                    & 40.35 \\ \cline{1-7}
\rowcolor[HTML]{FFFFFF} 
M4C        & Ms-OCR & \ding{52}          & 200M                         & {\color[HTML]{333333} 25.97}                   & {\color[HTML]{333333} 14.38}                    & 39.83 \\
\rowcolor[HTML]{EFEFEF} 
M5C-mbert  & Ms-OCR & \ding{52}          & 162M                        & {\color[HTML]{333333} 33.48}                   & {\color[HTML]{333333} 13.11}                    & 42.26 \\
\rowcolor[HTML]{FFFFFF} 
LaTr-base  & Ms-OCR & \ding{55}          & 226M                        & {\color[HTML]{333333} 36.47}                   & {\color[HTML]{333333} 20.25}                    & 39.83 \\
\rowcolor[HTML]{EFEFEF} 
mLaTr-base & Ms-OCR & \ding{55}          & 586M                        & {\color[HTML]{333333} \textbf{45}}                  & {\color[HTML]{333333}\textbf{44.3}}           & \textbf{46.06}\\
\rowcolor[HTML]{FFFFFF} 
LaTr-base  & Ms-OCR & \ding{52}          & 226M                        & {\color[HTML]{333333}37.08}                  & {\color[HTML]{333333}21.53}                   & 39.99 \\
\rowcolor[HTML]{EFEFEF} 
mLaTr-base & Ms-OCR & \ding{52}          & 586M                        & {\color[HTML]{333333} \textbf{45.01}}         & {\color[HTML]{333333}\textbf{44.25}}                   & \textbf{45.95}
\end{tabular}
\caption{\textbf{Results on the ML-TextVQA dataset}. Results refer to zero-shot transfer on Italian (IT) and Greek (EL) with multi-lingual models trained on English, Catalan, Spanish, and Chinese. Results are reported in term of Accuracy.}
\label{tab:ML_TextVQA-zero_shot}
\end{table}

%% file: tables/stvqa-iid.tex
\begin{table}[t!]
\centering
\tiny
\begin{tabular}{llccllllllll|ll}
                                   &                                & \multicolumn{1}{l}{}                             & \multicolumn{1}{l}{}                                & \multicolumn{2}{c}{EN}                                                                                           & \multicolumn{2}{c}{CA}                                                                             & \multicolumn{2}{c}{ES}                                                                             & \multicolumn{2}{c|}{ZH}                                                                             & \multicolumn{2}{c}{Avg}                                                                            \\ \cline{5-14} 
\multirow{-2}{*}{Method}           & \multirow{-2}{*}{OCR}          & \multicolumn{1}{l}{\multirow{-2}{*}{Vis. Feat.}} & \multicolumn{1}{l}{\multirow{-2}{*}{Params}}        & \multicolumn{1}{c}{\cellcolor[HTML]{FFFFFF}Acc}               & \multicolumn{1}{c}{\cellcolor[HTML]{FFFFFF}ANLS} & \multicolumn{1}{c}{\cellcolor[HTML]{FFFFFF}Acc} & \multicolumn{1}{c}{\cellcolor[HTML]{FFFFFF}ANLS} & \multicolumn{1}{c}{\cellcolor[HTML]{FFFFFF}Acc} & \multicolumn{1}{c}{\cellcolor[HTML]{FFFFFF}ANLS} & \multicolumn{1}{c}{\cellcolor[HTML]{FFFFFF}Acc} & \multicolumn{1}{c|}{\cellcolor[HTML]{FFFFFF}ANLS} & \multicolumn{1}{c}{\cellcolor[HTML]{FFFFFF}Acc} & \multicolumn{1}{c}{\cellcolor[HTML]{FFFFFF}ANLS} \\ \cline{5-14} 
\rowcolor[HTML]{FFFFFF} 
M4C                                & Ros-en                         &\ding{52}                                                & 200M                                                 & \cellcolor[HTML]{FFFFFF}{\color[HTML]{333333} 35.01}          & {\color[HTML]{333333} 0.439}                     & {\color[HTML]{333333} 34.74}                    & {\color[HTML]{333333} 0.438}                     & {\color[HTML]{333333} 34.36}                    & {\color[HTML]{333333} 0.435}                     & {\color[HTML]{333333} 30.4}                     & {\color[HTML]{333333} 0.384}                      & {\color[HTML]{333333} 33.63}                    & {\color[HTML]{333333} 0.424}                     \\
\rowcolor[HTML]{EFEFEF} 
\cellcolor[HTML]{EFEFEF}M5C-mbert  & \cellcolor[HTML]{EFEFEF}Ros-en & \cellcolor[HTML]{EFEFEF}\ding{52}                        & \cellcolor[HTML]{EFEFEF}{\color[HTML]{333333} 162M} & \cellcolor[HTML]{EFEFEF}{\color[HTML]{333333} 35.27}          & {\color[HTML]{333333} 0.438}                     & {\color[HTML]{333333} 35.27}                    & {\color[HTML]{333333} 0.438}                     & {\color[HTML]{333333} 35.81}                    & {\color[HTML]{333333} 0.444}                     & {\color[HTML]{333333} 35.24}                    & {\color[HTML]{333333} 0.438}                      & {\color[HTML]{333333} 35.4}                     & {\color[HTML]{333333} 0.439}                     \\
\rowcolor[HTML]{FFFFFF} 
LaTr-base                          & Ros-en                         &\ding{55}                                                & 226M                                                & \cellcolor[HTML]{FFFFFF}{\color[HTML]{333333} 41.59}          & {\color[HTML]{333333} 0.515}                     & {\color[HTML]{333333} 38.78}                    & {\color[HTML]{333333} 0.495}                     & {\color[HTML]{333333} 38.47}                    & {\color[HTML]{333333} 0.497}                     & {\color[HTML]{333333} 24.35}                    & {\color[HTML]{333333} 0.324}                      & {\color[HTML]{333333} 35.8}                     & {\color[HTML]{333333} 0.46}                      \\
\rowcolor[HTML]{EFEFEF} 
\cellcolor[HTML]{EFEFEF}mLaTr-base & \cellcolor[HTML]{EFEFEF}Ros-en & \cellcolor[HTML]{EFEFEF}\ding{55}                        & \cellcolor[HTML]{EFEFEF}586M                        & \cellcolor[HTML]{EFEFEF}{\color[HTML]{333333} 41.29}          & {\color[HTML]{333333} 0.526}                     & {\color[HTML]{333333} 41.29}                    & {\color[HTML]{333333} 0.522}                     & {\color[HTML]{333333} 41.44}                    & {\color[HTML]{333333} 0.528}                     & {\color[HTML]{333333} 40.07}                    & {\color[HTML]{333333} 0.507}                      & {\color[HTML]{333333} 41.03}                    & {\color[HTML]{333333} 0.521}                     \\
\rowcolor[HTML]{FFFFFF} 
LaTr-base                          & Ros-en                         &\ding{52}                                                & 226M                                                & \cellcolor[HTML]{FFFFFF}{\color[HTML]{333333} 41.67}          & {\color[HTML]{333333} 0.533}                     & {\color[HTML]{333333} 39.23}                    & {\color[HTML]{333333} 0.51}                      & {\color[HTML]{333333} 39}                       & {\color[HTML]{333333} 0.5}                       & {\color[HTML]{333333} 24.47}                    & {\color[HTML]{333333} 0.331}                      & {\color[HTML]{333333} 36.09}                    & {\color[HTML]{333333} 0.468}                     \\
\rowcolor[HTML]{EFEFEF} 
\cellcolor[HTML]{EFEFEF}mLaTr-base & \cellcolor[HTML]{EFEFEF}Ros-en & \cellcolor[HTML]{EFEFEF}\ding{52}                        & \cellcolor[HTML]{EFEFEF}586M                        & \cellcolor[HTML]{EFEFEF}{\color[HTML]{333333} 40.72}          & {\color[HTML]{333333} 0.518}                     & {\color[HTML]{333333} 40.68}                    & {\color[HTML]{333333} 0.517}                     & {\color[HTML]{333333} 40.45}                    & {\color[HTML]{333333} 0.514}                     & {\color[HTML]{333333} 39.5}                     & {\color[HTML]{333333} 0.504}                      & {\color[HTML]{333333} 40.33}                    & {\color[HTML]{333333} 0.513}                     \\ \hline
\rowcolor[HTML]{FFFFFF} 
M4C                                & Ms-OCR                         &\ding{52}                                                & 200M                                                 & \cellcolor[HTML]{FFFFFF}{\color[HTML]{333333} 41.9}           & {\color[HTML]{333333} 0.507}                     & {\color[HTML]{333333} 41.4}                     & {\color[HTML]{333333} 0.5}                       & {\color[HTML]{333333} 41.51}                    & {\color[HTML]{333333} 0.504}                     & {\color[HTML]{333333} 36.15}                    & {\color[HTML]{333333} 0.44}                       & {\color[HTML]{333333} 40.24}                    & {\color[HTML]{333333} 0.488}                     \\
\rowcolor[HTML]{EFEFEF} 
M5C-mbert                          & Ms-OCR                         &\ding{52}                                                & 162M                                                & {\color[HTML]{333333} 41.29}                                  & {\color[HTML]{333333} 0.505}                     & {\color[HTML]{333333} 42.39}                    & {\color[HTML]{333333} 0.518}                     & {\color[HTML]{333333} 42.16}                    & {\color[HTML]{333333} 0.514}                     & {\color[HTML]{333333} 41.74}                    & {\color[HTML]{333333} 0.509}                      & {\color[HTML]{333333} 41.89}                    & {\color[HTML]{333333} 0.512}                     \\
\rowcolor[HTML]{FFFFFF} 
LaTr-base                          & Ms-OCR                         & \ding{55}                                                & 226M                                                & \cellcolor[HTML]{FFFFFF}{\color[HTML]{333333} 47.07}          & {\color[HTML]{333333} 0.559}                     & {\color[HTML]{333333} 44.94}                    & {\color[HTML]{333333} 0.538}                     & {\color[HTML]{333333} 44.86}                    & {\color[HTML]{333333} 0.54}                      & {\color[HTML]{333333} 28.73}                    & {\color[HTML]{333333} 0.352}                      & {\color[HTML]{333333} 41.4}                     & {\color[HTML]{333333} 0.497}                     \\
\rowcolor[HTML]{EFEFEF} 
\cellcolor[HTML]{EFEFEF}mLaTr-base & \cellcolor[HTML]{EFEFEF}Ms-OCR & \cellcolor[HTML]{EFEFEF}\ding{55}                       & \cellcolor[HTML]{EFEFEF}586M                        & \cellcolor[HTML]{EFEFEF}{\color[HTML]{333333} 48.21}          & {\color[HTML]{333333} 0.572}                     & {\color[HTML]{333333} 47.72}                    & {\color[HTML]{333333} 0.568}                     & {\color[HTML]{333333} 47.53}                    & {\color[HTML]{333333} 0.566}                     & {\color[HTML]{333333} \textbf{47.07}}           & {\color[HTML]{333333} 0.555}                      & {\color[HTML]{333333} 47.63}                    & {\color[HTML]{333333} 0.565}                     \\
\rowcolor[HTML]{FFFFFF} 
LaTr-base                          & Ms-OCR                         &\ding{52}                                                & 226M                                                & \cellcolor[HTML]{FFFFFF}{\color[HTML]{333333} 47.34}          & {\color[HTML]{333333} 0.56}                      & {\color[HTML]{333333} 45.4}                     & {\color[HTML]{333333} 0.54}                      & {\color[HTML]{333333} 45.4}                     & {\color[HTML]{333333} 0.542}                     & {\color[HTML]{333333} 28.54}                    & {\color[HTML]{333333} 0.352}                      & {\color[HTML]{333333} 41.67}                    & {\color[HTML]{333333} 0.499}                     \\
\rowcolor[HTML]{EFEFEF} 
\cellcolor[HTML]{EFEFEF}mLaTr-base & \cellcolor[HTML]{EFEFEF}Ms-OCR & \cellcolor[HTML]{EFEFEF}\ding{52}                        & \cellcolor[HTML]{EFEFEF}586M                        & \cellcolor[HTML]{EFEFEF}{\color[HTML]{333333} \textbf{48.71}} & {\color[HTML]{333333} \textbf{0.583}}            & {\color[HTML]{333333} \textbf{47.91}}           & {\color[HTML]{333333} \textbf{0.574}}            & {\color[HTML]{333333} \textbf{48.36}}           & {\color[HTML]{333333} \textbf{0.577}}            & {\color[HTML]{333333} 46.84}                    & {\color[HTML]{333333} \textbf{0.563}}             & {\color[HTML]{333333} \textbf{47.96}}           & {\color[HTML]{333333} \textbf{0.574}}           
\end{tabular}
\caption{\textbf{Results on the ML-STVQA dataset}. Results refer to multi-lingual training on English, Catalan, Spanish, and Chinese and are reported in term of Accuracy and ANLS \cite{biten2019scene}. Microsoft-OCR improve from 5\% to 10\% over all methods. Visual features do not increase accuracy in general.}
\label{tab:ML_STVQA-iid}
\end{table}

%% file: tables/stvqa-zero_shot.tex
\begin{table}[t!]
\begin{tabular}{llcccccc|cc}
\rowcolor[HTML]{FFFFFF} 
\cellcolor[HTML]{FFFFFF}                         & \cellcolor[HTML]{FFFFFF}                      & \multicolumn{1}{l}{\cellcolor[HTML]{FFFFFF}}                           & \multicolumn{1}{l}{\cellcolor[HTML]{FFFFFF}}                         & \multicolumn{2}{c}{\cellcolor[HTML]{FFFFFF}IT} & \multicolumn{2}{c|}{\cellcolor[HTML]{FFFFFF}EL} & \multicolumn{2}{l}{\cellcolor[HTML]{FFFFFF}Avg} \\ \cline{5-10} 
\rowcolor[HTML]{FFFFFF} 
\multirow{-2}{*}{\cellcolor[HTML]{FFFFFF}Method} & \multirow{-2}{*}{\cellcolor[HTML]{FFFFFF}OCR} & \multicolumn{1}{l}{\multirow{-2}{*}{\cellcolor[HTML]{FFFFFF}Vis.Feat}} & \multicolumn{1}{l}{\multirow{-2}{*}{\cellcolor[HTML]{FFFFFF}Params}} & acc                    & ANLS                  & acc                    & ANLS                   & acc                    & ANLS                   \\ \hline
\rowcolor[HTML]{FFFFFF} 
M4C                                              & Ros-en                                        & \ding{52}                                                                      & 200M                                                                 & 29.15                  & 0.357                 & 21.77                  & 0.288                  & 33.63                  & 0.424                  \\
\rowcolor[HTML]{EFEFEF} 
M5C-mbert                                        & Ros-en                                        & \ding{52}                                                                      & 162M                                                                 & 30.94                  & 0.389                 & 24.58                  & 0.306                  & 35.4                   & 0.439                  \\
\rowcolor[HTML]{FFFFFF} 
LaTr-base                                        & Ros-en                                        & \ding{55}                                                                      & 226M                                                                 & 34.78                  & 0.451                 & 23.1                   & 0.307                  & 35.8                   & 0.46                   \\
\rowcolor[HTML]{EFEFEF} 
mLaTr-base                                       & Ros-en                                        & \ding{55}                                                                      & 586M                                                                 & 39.8                   & 0.505                 & 38.55                  & 0.494                  & 41.03                  & 0.521                  \\
\rowcolor[HTML]{FFFFFF} 
LaTr-base                                        & Ros-en                                        & \ding{52}                                                                      & 226M                                                                 & 34.89                  & 0.453                 & 24.05                  & 0.324                  & 36.09                  & 0.468                  \\
\rowcolor[HTML]{EFEFEF} 
mLaTr-base                                       & Ros-en                                        & \ding{52}                                                                      & 586M                                                                 & 39.04                  & 0.501                 & 38.13                  & 0.485                  & 40.33                  & 0.513                  \\ \hline
\rowcolor[HTML]{FFFFFF} 
M4C                                              & Ms-OCR                                        & \ding{52}                                                                      & 200M                                                                 & 34.02                  & 0.413                 & 23.4                   & 0.293                  & 40.24                  & 0.488                  \\
\rowcolor[HTML]{EFEFEF} 
M5C-mbert                                        & Ms-OCR                                        & \ding{52}                                                                      & 162M                                                                 & 38.58                  & 0.468                 & 30.78                  & 0.384                  & 41.89                  & 0.512                  \\
\rowcolor[HTML]{FFFFFF} 
LaTr-base                                        & Ms-OCR                                        & \ding{55}                                                                      & 226M                                                                 & 40.6                   & 0.486                 & 27.17                  & 0.329                  & 41.4                   & 0.497                  \\
\rowcolor[HTML]{EFEFEF} 
mLaTr-base                                       & Ms-OCR                                        & \ding{55}                                                                      & 586M                                                                 & \textbf{46.54}         & \textbf{0.557}        & \textbf{45.97}         & \textbf{0.546}         & \textbf{47.63}         & \textbf{0.565}         \\
\rowcolor[HTML]{FFFFFF} 
LaTr-base                                        & Ms-OCR                                        & \ding{52}                                                                      & 226M                                                                 & 40.72                  & 0.489                 & 28.16                  & 0.347                  & 41.67                  & 0.498                  \\
\rowcolor[HTML]{EFEFEF} 
mLaTr-base                                       & Ms-OCR                                        & \ding{52}                                                                      & 586M                                                                 & 46.35                  & 0.554                 & 44.75                  & 0.538                  & \textbf{47.96}         & \textbf{0.574}        
\end{tabular}
\caption{\textbf{Results on the ML-STVQA dataset}. Results refer to zero-shot transfer on Italian (IT) and Greek (EL) with multi-lingual models trained on English, Catalan, Spanish, and Chinese. Results are reported in term of Accuracy and ANLS (cite ANLS).}
\label{tab:ML_STVQA-zero_shot}
\end{table}


%% file: tables/textvqa_translate_robustness-tabular.tex
\begin{tabular}{lccc|cc|c}
         & \cellcolor[HTML]{FFFFFF}CA & \cellcolor[HTML]{FFFFFF}ES & \cellcolor[HTML]{FFFFFF}ZH & \cellcolor[HTML]{FFFFFF}IT & \cellcolor[HTML]{FFFFFF}EL & \cellcolor[HTML]{FFFFFF}\textbf{Avg} \\ \hline
\rowcolor[HTML]{FFFFFF} 
OPUS     & 42.25                      & 45.73                      & 43.82                      & 44.53                      & 43.39                      & \cellcolor[HTML]{FFFFFF}\textbf{46.06}     \\
\rowcolor[HTML]{EFEFEF} 
M2M\_100 & 45.73                      & 45.69                      & 44.39                      & 44.91                      & 43.29                      & \textbf{46.06}                             \\
\rowcolor[HTML]{FFFFFF} 
mBART    & /                          & 45.76                      & 43.53                      & 44.81                      & /                          & \cellcolor[HTML]{FFFFFF}\textbf{46.06}    
\end{tabular}

%% file: tables/stvqa_translate_robustness-tabular.tex
\begin{tabular}{
>{\columncolor[HTML]{FFFFFF}}l 
>{\columncolor[HTML]{FFFFFF}}l 
>{\columncolor[HTML]{FFFFFF}}l 
>{\columncolor[HTML]{FFFFFF}}l |
>{\columncolor[HTML]{FFFFFF}}l 
>{\columncolor[HTML]{FFFFFF}}l |l}
                                 & CA                            & ES                            & ZH                            & IT                            & EL                            & \textbf{Avg}   \\ \hline
OPUS                             & 46.84                         & 47.72                         & 45.74                         & 46.31                         & 46.84                         & \textbf{47.96} \\
\cellcolor[HTML]{EFEFEF}M2M\_100 & \cellcolor[HTML]{EFEFEF}47.22 & \cellcolor[HTML]{EFEFEF}47.68 & \cellcolor[HTML]{EFEFEF}45.97 & \cellcolor[HTML]{EFEFEF}46.16 & \cellcolor[HTML]{EFEFEF}45.09 & \textbf{47.96} \\
mBART                            & -                             & 46.96                         & 45.89                         & 45.93                         & -                             & \textbf{47.96}
\end{tabular}